%% file: main.tex
\documentclass[final]{cvpr}

\usepackage{times}
\usepackage{epsfig}
\usepackage{graphicx}
\usepackage{amsmath}
\usepackage{amssymb}

\usepackage[export]{adjustbox}
\usepackage{booktabs}
\usepackage{multirow}
\usepackage{adjustbox}
\usepackage{caption}
\usepackage{cite}                      
\usepackage{stmaryrd}
\usepackage{pifont}
\usepackage{subcaption}

\usepackage[pagebackref=true,breaklinks=true,colorlinks,bookmarks=false]{hyperref}

\def\cvprPaperID{2066} 
\def\confYear{CVPR 2021}

\newcommand*\colorcmark[1]{%
  \expandafter\newcommand\csname #1cmark\endcsname{\textcolor{#1}{\ding{51}}}%
}
\colorcmark{green}

\newcommand*\colorxmark[1]{%
  \expandafter\newcommand\csname #1xmark\endcsname{\textcolor{#1}{\ding{55}}}%
}
\colorxmark{red}

\begin{document}

\title{Beyond Short Clips: End-to-End Video-Level Learning\\with Collaborative Memories}

\author{Xitong Yang\textsuperscript{1}\thanks{Work done during an internship at Facebook AI.}, Haoqi Fan\textsuperscript{2}, Lorenzo Torresani\textsuperscript{2,3}, Larry Davis\textsuperscript{1}, Heng Wang\textsuperscript{2}\\
\textsuperscript{1}University of Maryland, College Park~ \textsuperscript{2}Facebook AI~ \textsuperscript{3}Dartmouth\\
{\tt\small \{xyang35,lsd\}@cs.umd.edu~ \{haoqifan,torresani,hengwang\}@fb.com}}
	

\maketitle
\thispagestyle{empty}    

\makeatletter
\DeclareRobustCommand\onedot{\futurelet\@let@token\@onedot}
\def\@onedot{\ifx\@let@token.\else.\null\fi\xspace}

\def\eg{\emph{e.g}\onedot} \def\Eg{\emph{E.g}\onedot}
\def\ie{\emph{i.e}\onedot} \def\Ie{\emph{I.e}\onedot}
\def\cf{\emph{c.f}\onedot} \def\Cf{\emph{C.f}\onedot}
\def\etc{\emph{etc}\onedot} \def\vs{\emph{vs}\onedot}
\def\wrt{w.r.t\onedot} \def\dof{d.o.f\onedot}
\def\etal{\emph{et al}\onedot}
\makeatother

\def\cm{\textsc{CM}\xspace}

\newcommand{\hq}[1]{{\color{blue}haoqi: #1}}
\newcommand{\xt}[1]{{\color{cyan}Xitong: #1}}
\newcommand{\hw}[1]{{\color{green}Heng: #1}}
\newcommand{\LT}[1]{{\color{red}LT: #1}}

\newcommand{\PLH}{{\mkern-2mu\times\mkern-2mu}}

\newcommand{\sota}{state-of-the-art\xspace}


\input{abs}
\input{intro}

\input{related}

\input{approach}

\input{setup}

\input{exp}
\input{conclusion}

{\small
\bibliographystyle{ieee_fullname}
\bibliography{egbib}
}

\clearpage
\input{appendix}

\end{document}

%% file: abs.tex
\begin{abstract}

The standard way of training video models entails sampling at each iteration a single clip from a video and optimizing the clip prediction with respect to the video-level label. We argue that a single clip may not have enough temporal coverage to exhibit the label to recognize, since video datasets are often weakly labeled with categorical information but without dense temporal annotations. Furthermore, optimizing the model over brief clips impedes its ability to learn long-term temporal dependencies. To overcome these limitations, we introduce a collaborative memory mechanism that encodes information across multiple sampled clips of a video at each training  iteration. This enables the learning of long-range dependencies beyond a single clip. We explore different design choices for the collaborative memory to ease the optimization difficulties. Our proposed framework is end-to-end trainable  and  significantly improves the accuracy of video classification at a negligible computational overhead. Through extensive experiments, we demonstrate that our framework generalizes to different video architectures and tasks, outperforming the state of the art on both action recognition (\eg, Kinetics-400 \& 700, Charades, Something-Something-V1) and action detection (\eg, AVA v2.1 \& v2.2).

\end{abstract}

%% file: intro.tex
\vspace{-0.1in}
\section{Introduction}\label{sec:intro}

\begin{figure}[t]
    \centering
    \includegraphics[width=0.9\linewidth]{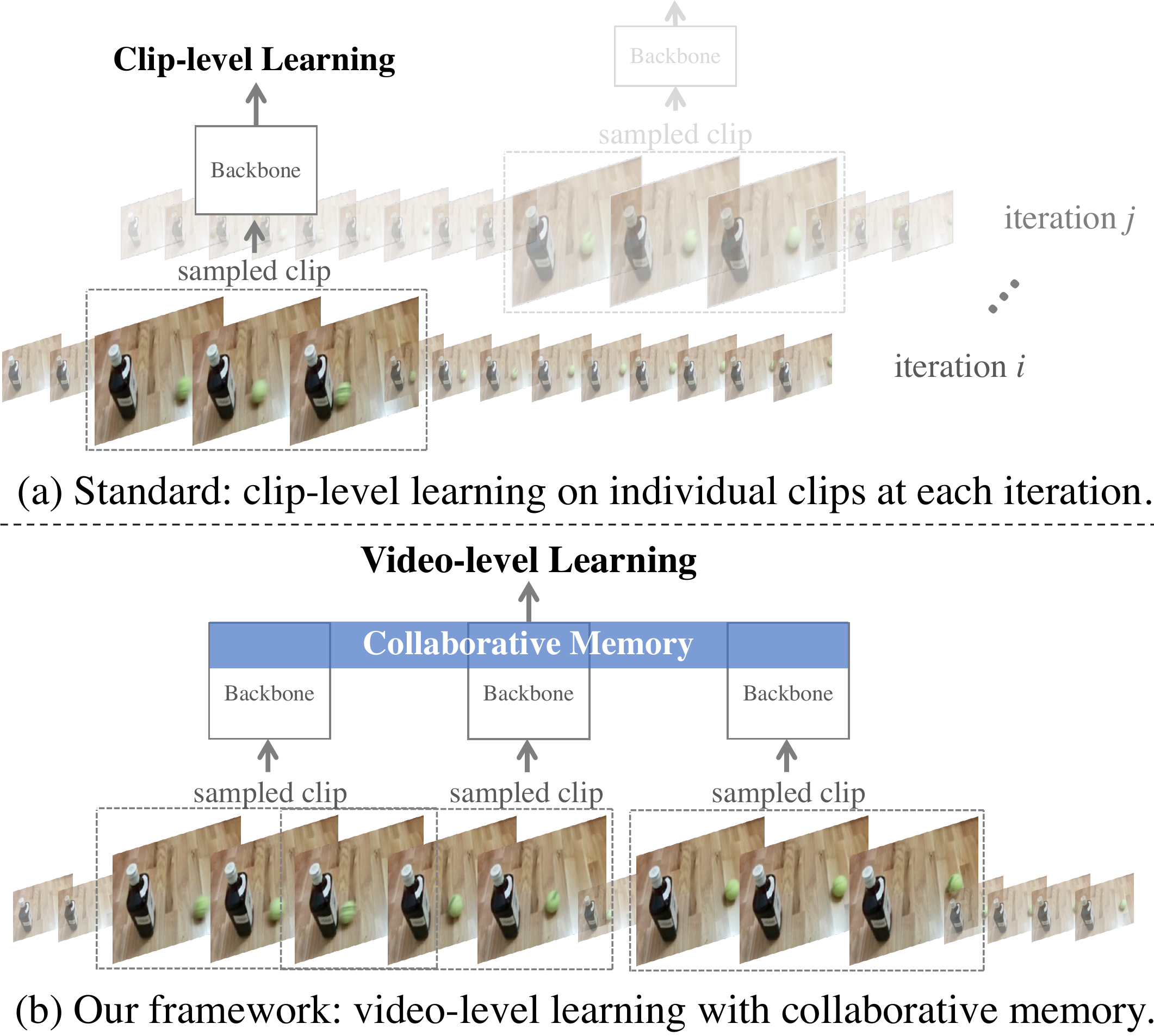}
    \vspace{-0.05in}
    \caption{Clip-level learning \vs our proposed end-to-end video-level learning framework. (Action label: \textit{something being deflected from something}.)
    }
    \label{fig:teaser}
    \vspace{-0.15in}
\end{figure}


In recent years, end-to-end learning of 3D convolutional networks (3D CNNs) has emerged as the prominent paradigm for video classification~\cite{baccouche2011sequential,Ming2013,Tran15,linsuniccv15,tran2017closer,P3D,I3D,xie2017rethinking,wang2017non,diba2019dynamonet,feichtenhofer2020x3d,wang2020video,feichtenhofer2019slowfast,tran2019video}. 
Steady improvements in accuracy have come with the introduction of increasingly deeper and larger networks. However, due to their high computational cost and large memory requirements, most video models are optimized at each iteration over short, fixed-length clips rather than the entire video.  

Although widely used in modern video models, the clip-level learning framework is sub-optimal for video-level classification. First, capturing long-range temporal structure beyond short clips is not possible as the models are only exposed to individual clips during training. Second, the video-level label may not be well represented in a brief clip, which may be an uninformative segment of the video or include only a portion of the action, as shown in Figure~\ref{fig:teaser}(a). Thus, optimizing a model over individual clips using video-level labels is akin to training with noisy labels.
Recent attempts to overcome these limitations include methods that build a separate network on top of the clip-based backbone~\cite{wu2019long,zhu2019faster,hussein2019timeception}.
However, these approaches either cannot be trained end-to-end with the backbone (\ie, the video model is optimized over pre-extracted clip-level features) or require ad-hoc backbones 
which hinder their application in the current landscape of evolving architectures.


In this paper, we propose an end-to-end learning framework that optimizes the classification model using video-level information collected from multiple temporal locations of the video, shown in Figure~\ref{fig:teaser}(b). Our approach hinges on a collaborative memory mechanism that accumulates video-level contextual information from multiple clips sampled from the video. Within the same training iteration, this contextual information is shared back with all the clips to enhance the individual clip representations. The collaborative memory allows the model to capture long-range temporal dependencies beyond individual short clips by generating clip-specific memories that encode the relation between each local clip and the global video-level context.

Our experiments demonstrate that the proposed training framework is effective and generic. Specifically, our approach does not make any assumption about the backbone architecture. We empirically show that it consistently yields significant gains in accuracy when applied to different state-of-the-art architectures (e.g. SlowFast~\cite{feichtenhofer2019slowfast}, R(2+1)D~\cite{tran2017closer}, I3D-NL~\cite{wang2017non}). We also introduce and compare several design variants of the collaborative memory. Furthermore, we demonstrate that the accuracy improvements come at a negligible computational overhead and without an increase in memory requirements. Finally, we show that our framework can be extended to action detection where it yields significant improvements without requiring extra information, such as optical flow and object detection predictions, which are commonly used in previous  work~\cite{Sun2018ACRN,tang2020asynchronous}.
We summarize our major contributions as follows:


\begin{itemize} 
\setlength\itemsep{0.1em}
\item A new framework that enables end-to-end learning of video-level dependencies for clip-based models.
\item A new collaborative memory mechanism that facilitates information exchange across multiple clips. We explore different design choices 
and provide insights about the optimization difficulties.
\item Experiments demonstrating that our collaborative memory framework generalizes to different backbones and tasks, producing state of the art results for action recognition and detection.
\end{itemize}

%% file: related.tex
\section{Related Work}\label{sec:related}

\noindent\textbf{Clip-Level Video Architectures.}
Since the introduction of 3D CNNs~\cite{baccouche2011sequential,Ming2013,Tran15} to video classification, new architectures~\cite{linsuniccv15,tran2017closer,P3D,I3D,xie2017rethinking,wang2017non,diba2019dynamonet,feichtenhofer2020x3d,wang2020video} have been proposed to learn better spatiotemporal representations. Besides models aimed at improving accuracy, several architectures have been proposed to achieve better performance/cost trade-offs~\cite{tran2019video,feichtenhofer2020x3d,tinynetworks,zolfaghari2018eco,kopuklu2019resource,lin2019tsm,9216180}. 
Another line of research involves the design of multiple-stream networks~\cite{SimonyanZ14,yue2015beyond,WangXW0LTG16,FeichtenhoferPZ16,I3D,xie2017rethinking,feichtenhofer2019slowfast}, where each stream consumes a different type of input, \eg, RGB or optical flow. Besides CNNs, transformer-based models, \eg, TimeSformer~\cite{bertasius2021space}, also show promising results. Unlike prior work focused on the design of clip-level architecture, our paper proposes a new framework to learn long-range dependencies using existing clip-level models. As we do not make any assumption about the clip-level architecture, our framework generalizes to different backbones and enables end-to-end training of clip models with video-level contextual information. 

\vspace{-0.17in}
\paragraph{Video-Level Classification.} 
Several attempts have been made to overcome the limitations of the single-clip training framework. Timeception~\cite{hussein2019timeception} uses multi-scale temporal convolutions to cover different temporal extents for long-range temporal modeling. Timeception layers are trained on top of a frozen backbone. 
Both TSN~\cite{WangXW0LTG16} and ECO~\cite{zolfaghari2018eco} divide the input video into segments of equal size and randomly sample a short snippet or a single frame from each segment to provide better temporal coverage during training. As the GPU memory cost grows linearly \wrt the number of segments, TSN and ECO adopt lightweight 3D CNNs or even 2D CNNs as backbones in order to process multiple segments simultaneously. These simple backbones limit the performance of the framework. In addition, TSN uses averaging to aggregate the predictions from different segments, whereas we propose a dedicated memory mechanism to model the video-level context. 
FASTER~\cite{zhu2019faster} and SCSampler~\cite{korbar2019scsampler} explore strategies to limit the detrimental impact of applying video-level labels to clips and to save computational cost. 

Another related work to our approach is LFB~\cite{wu2019long}. It leverages context features from other clips to augment the prediction on the current clip. Unlike our approach, context features stored in LFB are {\em pre-computed} using a {\em separate} model. As a result, the context features cannot be updated during the training and the model used to extract these context features is not optimized for the task. In contrast, our framework is end-to-end trainable and the accumulated contextual information can back-propagate into the backbone architecture. Note that storing context features is infeasible for large-scale video datasets, \eg, Kinetics, and LFB is mainly designed for action detection applications.

\vspace{-0.17in}
\paragraph{Learning With Memories.} 
Memory mechanisms~\cite{weston2014memory,sukhbaatar2015end,grave2016improving,ba2016using} have been widely used in Recurrent Neural Networks for language modeling in order to learn long-term dependencies from sequential text data.
Specifically, memory networks~\cite{weston2014memory} have been proposed for question answering (QA), while Sukhbaatar \etal~\cite{sukhbaatar2015end} have introduced a strategy enabling end-to-end learning of these models. RWMN~\cite{na2017read} has extended the QA application on movie videos. Grave \etal ~\cite{grave2016improving} have proposed to store past hidden activations as a memory that can be accessed through a dot product with the current hidden activation. 

These works are similar in spirit to our approach, but our application is in a different domain with different constraints and challenges.
Moreover, our collaborative memory mechanism is designed to capture the interactions among the samples, is extremely lightweight and memory-friendly, and is suitable for training computationally intensive video models.



%% file: approach.tex
\section{End-to-End Video-Level Learning with Collaborative Memory}\label{sec:approach}

We start with an overview of the proposed framework, then present different designs for the collaborative memory. We conclude with a discussion of implementation strategies to cope with the GPU memory constraint.


\subsection{Overview of the Proposed Framework}

Given a video recognition architecture (\eg, I3D~\cite{I3D}, R(2+1)D~\cite{tran2017closer}, SlowFast~\cite{feichtenhofer2019slowfast}) that operates on short, fixed-length clips, our goal is to perform video-level learning in an end-to-end manner.
In particular, we aim to optimize the \textit{clip-based} model using \textit{video-level} information collected from the whole video.
To achieve this, we start by sampling multiple clips from the video within the same training iteration in order to cover different temporal locations of the video.
The clip-based representations generated from multiple clips are then accumulated via a collaborative memory mechanism that captures interactions among the clips and builds video-level contextual information.
After that, clip-specific memories are generated to enhance the individual clip-based representations by infusing the video-level information into the backbone.
Finally, the sampled clips are jointly optimized with a video-level supervision applied to the consensus of predictions from multiple clips.

\vspace{-0.13in}
\paragraph{Multi-clip sampling.} 
Given a video $\mathcal{V} = \left\{I_0, ..., I_{T-1} \right\}$ with $T$ frames, we sample $N$ clips $\{\mathcal{C}_{0}, ..., \mathcal{C}_{N-1}\}$ from the video at each training iteration. Each short clip $\mathcal{C}_{n} = \{I_{t_n}, ..., I_{t_n+L-1} \}$ consists of $L$ consecutive frames randomly sampled from the full-length video where $t_n$ indicates the index of the start frame. $N$ is a hyper-parameter that can be decided based on the ratio between the video length and the clip length to ensure sufficient temporal coverage. 
The sampled clips are then fed to the backbone to generate clip-based representations $\{X_n\}_{n=0}^{N-1}$, where $X_n = f(\mathcal{C}_n)$, and $f$ represents the clip-level backbone. In the traditional clip-level classification, $X_n$ 
is directly used to perform the final prediction via a classifier $h$ : ${\bf y}_n = h(X_n) =  h(f(\mathcal{C}_n))$, where ${\bf y}_n$ is the vector of classification scores.

\begin{figure}[t]
    \centering
    \includegraphics[width=0.9\linewidth]{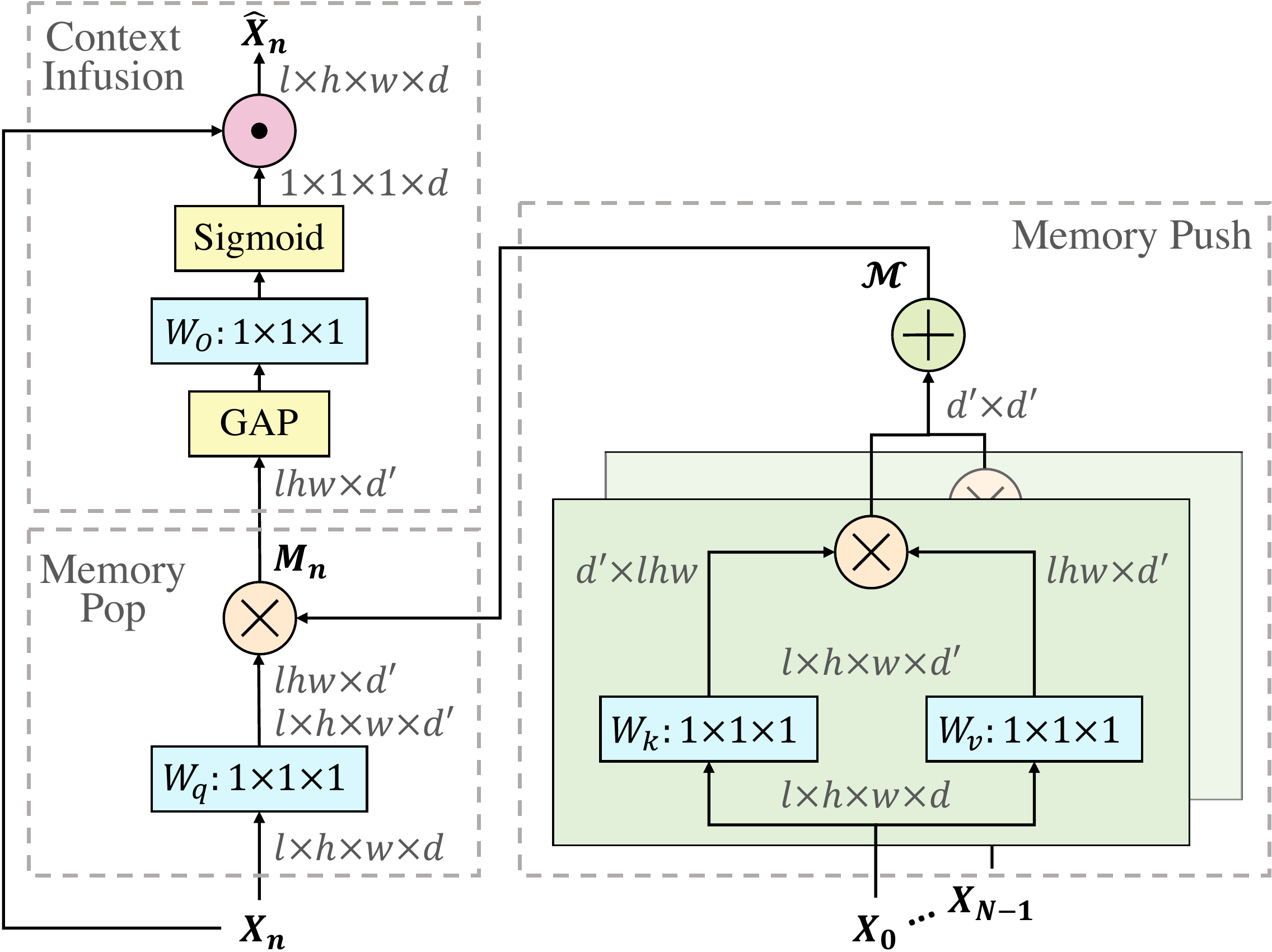}
    \caption{Collaborative memory with associative memory and feature gating. The feature maps are shown as the shape of their tensors, \eg, $l\times h\times w \times d$. GAP denotes global average pooling. $\otimes$ and $\odot$ indicate matrix and elementwise multiplication, respectively.}
    \label{fig:design}
    \vspace{-0.2in}
\end{figure}

\vspace{-0.13in}
\paragraph{Collaborative memory.}
Our approach hinges on a collaborative memory mechanism that accumulates information from multiple clips for learning video-level dependencies and then shares this video-level context back with the individual clips, as illustrated in Figure~\ref{fig:design}. Specifically, the collaborative memory involves two stages: 
\vspace{-0.05in}
\begin{itemize}
\setlength\itemsep{0.05em}
    \item \textit{Memory interactions: } A global memory of the whole video is constructed by accumulating information from multiple clips: $\mathcal{M} =  Push(\{X_n\}_{n=0}^{N-1})$. The global memory is then shared back with the individual clips in order to generate clip-specific memories: $M_n = Pop(\mathcal{M}, X_n)$.
    \item \textit{Context infusion: } The individual clip-based representations are infused with video-level context. This is done by means of a gating function $g$ that enhances each clip representation with the information stored in the clip-specific memory: $\hat{X}_n = g(X_n, M_n)$. 
\end{itemize}   
    

\vspace{-0.17in}
\paragraph{Video-level supervision.} To facilitate the joint optimization over multiple clips, we apply a video-level loss that takes into account the clip-level predictions as well as the video-level prediction aggregated from all $N$ sampled clips. Formally, we first aggregate the clip-level predictions via average pooling over $N$ clips: $H=\frac{1}{N}\sum_{n=0}^{N-1}h(\hat{X}_n)=\frac{1}{N}\sum_{n=0}^{N-1}h(g(f(\mathcal{C}_n), M_n))$. Then our video-level loss can be written as
\vspace{-0.05in}
\begin{equation}
\label{eq:loss}
    \mathcal{L}_{video} = \frac{1}{N}\sum_{n=0}^{N-1}\mathcal{L}(h(\hat{X}_n)) + \alpha \mathcal{L}(H).
    \vspace{-0.05in}
\end{equation}
$\mathcal{L}$ denotes the cross-entropy loss for classification and $\alpha$ is the weight to balance 
the two terms which account for the clip-level losses and the video-level loss aggregated from all $N$ clips. All the parameters 
(\ie, $f$, $g$, and $h$) are optimized end-to-end \wrt this objective.

\subsection{Collaborative memory}\label{sec:memory}

Our idea of collaborative memory is generic and can be implemented in a variety of ways. 
In this section we introduce a few possible designs. We empirically evaluate these different options in Section~\ref{sec:ablation}.

\vspace{-0.17in}
\paragraph{Memory interactions.}
The design of memory interactions should follow two principles: 1) The memory footprint for storing the global memory should be manageable; 2) Interactions with the memory should be computationally efficient. The first principle implies that the memory consumption should not grow with the number of clips $N$. Thus, simply storing all clip-based features is not feasible.

\textbf{-- Average Pooling: } Let the clip-based representation $X_n$ be a $k\times d$ matrix, where $k$ is the spatial-temporal resolution (\ie, $height \times width \times length$) 
and $d$ is the number of channels. A simple strategy is to perform a global average pooling over all sampled clips: $\mathcal{M} = Push(\{X_n\}_{n=0}^{N-1})= Pool(\{X_n W_I\}_{n=0}^{N-1})$.  $W_I\in\mathbb{R}^{d\times d'}$ is a learnable weight matrix to reduce the dimensionality from $d$ to $d'$. This global information can be simply shared back with all the clips: $M_n = Pop(\mathcal{M}, X_n) = \mathcal{M}$.

\textbf{-- Associative Memory: } 
Although avg/max pooling is capable of collecting information from multiple clips, it fails to capture the inter-clip dependencies and the clip-specific information cannot be retrieved from the global memory $\mathcal{M}$.
This motivates us to design a new mechanism that enables the retrieval of clip-specific information when needed.
Inspired by associative networks~\cite{ba2016using,Hopfield1982neural}, we propose to accumulate the clip-level features using the outer product operator to generate the global memory $\mathcal{M}$:
\vspace{-0.05in}
\begin{equation}
\label{eq:push}
   Push(\{X_n\}_{n=0}^{N-1}) = \frac{1}{N}\sum_{n=0}^{N-1} (X_n W_k)^T (X_n W_v).
\vspace{-0.05in}
\end{equation}
Given the $n$-th clip, we obtain its clip-specific memory by: 
\vspace{-0.05in}
\begin{equation}
\label{eq:pop}
   M_n = Pop(\mathcal{M}, X_n) = (X_n W_q)\mathcal{M},
\vspace{-0.05in}
\end{equation}
where $W_k, W_v, W_q\in\mathbb{R}^{d\times d'}$ are learnable weight matrices for memory interactions and dimension reduction.
Note that this memory design can be viewed as implementing a form of video-level inter-clip attention, where the clip-based representation $X_n$ attends to features generated from all sampled clips of the video in proportion to their similarities: $M_n = \frac{1}{N}\sum_{m=0}^{N-1} \left [(X_n W_q)(X_m W_k)^T \right ] (X_m W_v)$. However, unlike the self-attention mechanism ~\cite{wang2017non,Ashish2017attention}, our design is more efficient in both computation and memory consumption as it does not require to store all clip-level features or perform pairwise comparison between all the clips. 

\vspace{-0.17in}
\paragraph{Context Infusion.}
One way to incorporate the clip-specific memory $M_n$ with the clip-level features $X_n$ is through a residual connection: $\hat{X}_n = M_n W_O + X_n$, where $W_O \in \mathbb{R}^{d'\times d}$ is a linear transformation to match the feature dimensionality.
However, as we will show in our experiments (Figure~\ref{fig:curves}), this design tends to overfit to the clip-specific memory during training and leads to inferior performance.
As $M_n$ stores much more information 
than a single clip-level feature $X_n$, the model mostly relies on $M_n$ during training and makes little use of $X_n$.

In light of the above observation, we propose to infuse context information into the clip-level features through a \textit{feature gating} operation. Rather than allowing the model to directly access the clip-specific memory $M_n$, feature gating forces the model to recalibrate the strengths of different 
clip-level features using the contextual information.
Formally, the enhanced features are computed as
\begin{equation}
    \hat{X}_n = \left( J + \sigma(\hat{M}_n W_O)\right) \odot X_n,
\end{equation}
where $\sigma$ is the sigmoid activation function, $\odot$ is the elementwise multiplication and $J$ is an all-ones matrix for residual connection.
$\hat{M}_n$ is obtained by averaging the spatial and temporal dimensions of $M_n$: $\hat{M}_n = GAP(M_n)$. 
Our feature gating operation can be considered as a channel-wise attention mechanism similar to context gating~\cite{xie2017rethinking,miech2017learnable} and the SE block~\cite{hu2018squeeze}.
However, the attention weights in our method are generated by video-level contextual information, instead of self-gating values that capture channel-wise relationships within the same clip.
Experimental results show that our proposed feature gating design alleviates the optimization difficulties during training and enables a more effective use of the  video-level contextual information. 

\begin{figure*}[t!]
    \begin{minipage}[b]{0.3\linewidth}
        \centering
        \includegraphics[width=0.95\linewidth]{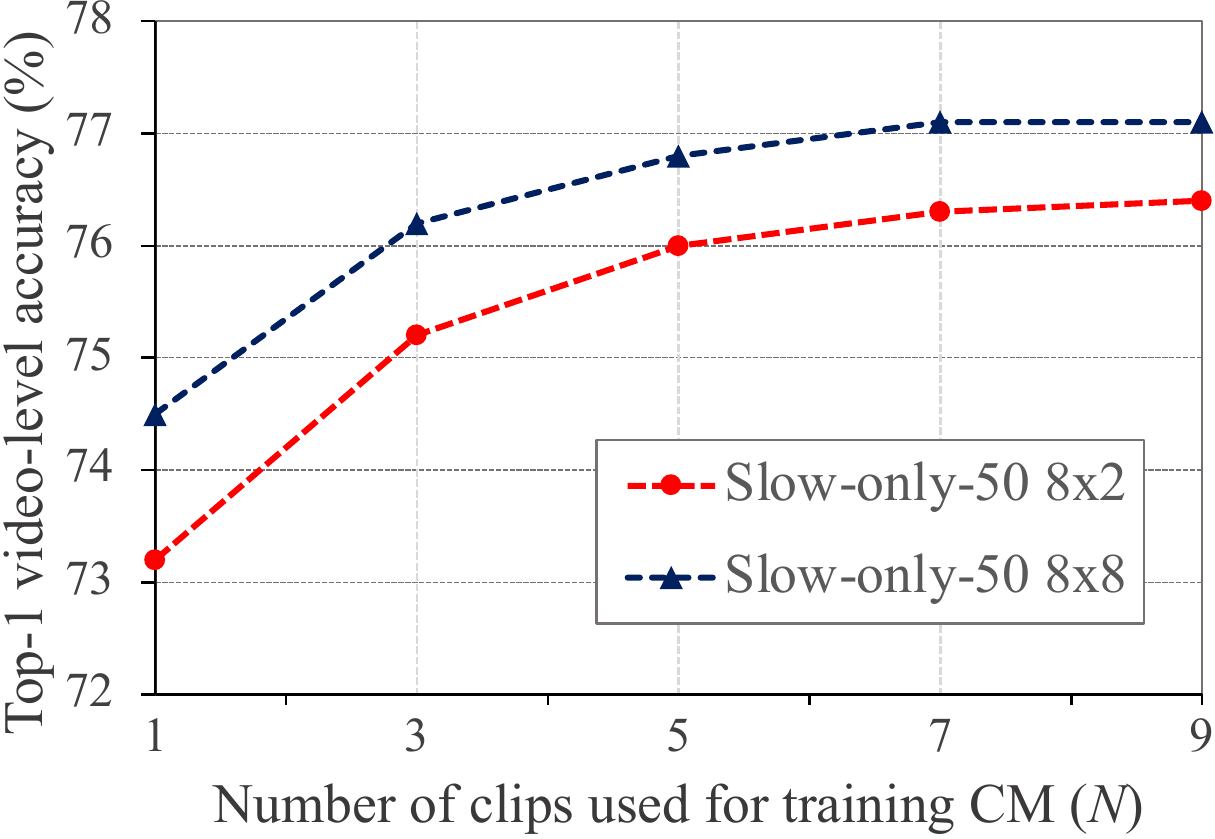}
        \vspace{-0.1in}
        \caption{\textbf{Video-level accuracy} on Kinetics-400 with CM. The horizontal axis shows the number of clips ($N$) used for training \cm. Note that all models use $30$ crops for inference.}
        \label{fig:main}
    \end{minipage}
    \hfill
    \begin{minipage}[b]{0.3\linewidth}
        \centering
        \includegraphics[width=0.95\linewidth]{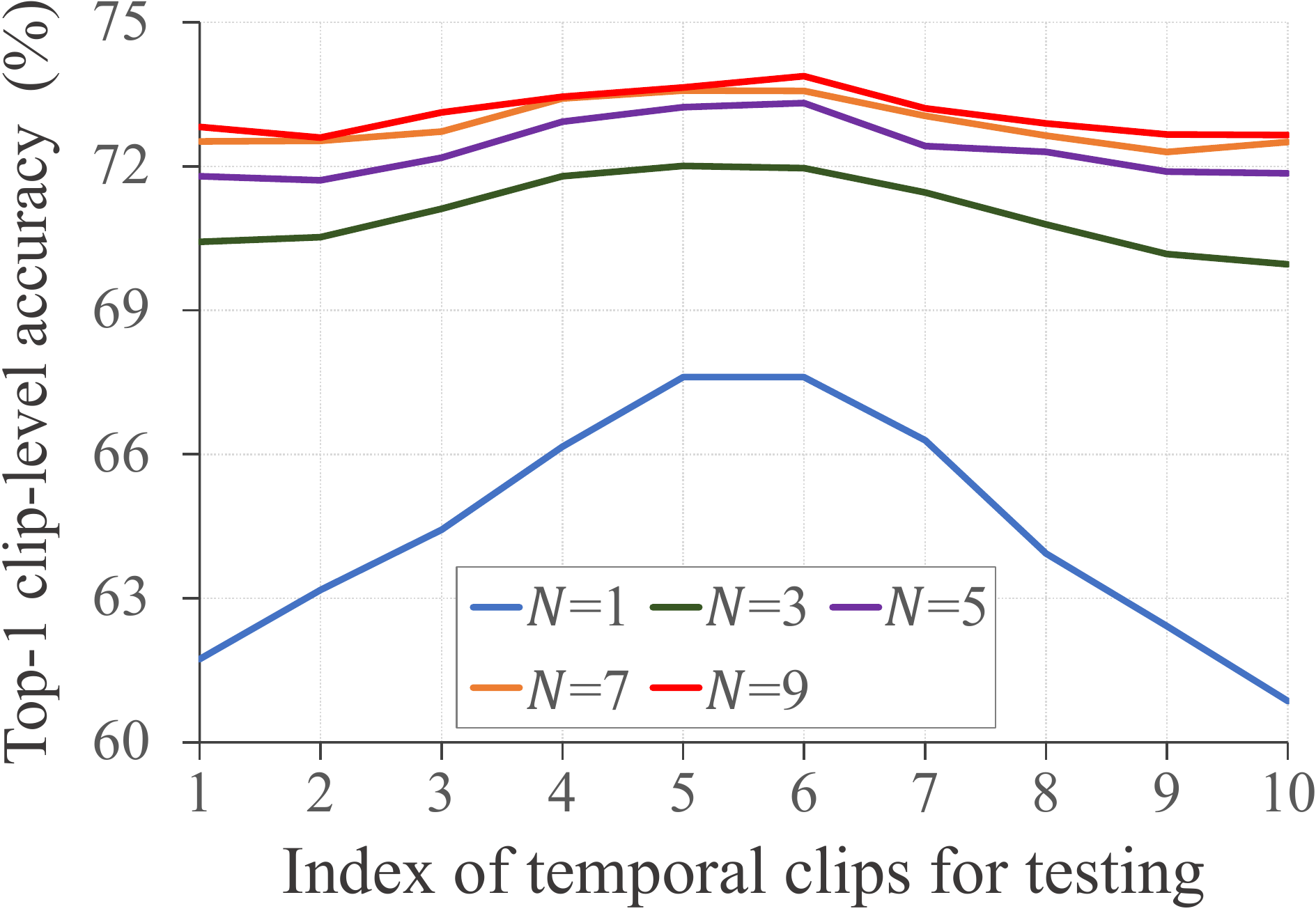}
        \vspace{-0.1in}
        \caption{\textbf{Clip-level accuracy} on Kinetics-400 at 10 different temporal locations within the video. 
        $N$ indicates the number of clips used for training \cm.}
        \label{fig:clip}
    \end{minipage}
    \hfill
    \begin{minipage}[b]{0.37\linewidth}
        \centering
        \footnotesize
        \renewcommand{\arraystretch}{1.2}
        \setlength{\tabcolsep}{0pt} 
        \begin{tabular*}{\linewidth}{@{\extracolsep{\fill}\,}lcccc@{}}
            \toprule
            \multicolumn{1}{c}{\bf Model} & \textbf{Baseline}&  \textbf{Ours} & \textbf{$\triangle$} & \textbf{FLOPs} \\
            \midrule
            Slow-only-50 8$\times$8~\cite{feichtenhofer2019slowfast} & 74.4 & \textbf{76.8} & +2.4 & 1.03$\times$\\
            I3D-50+NL 32$\times$2~\cite{wang2017non} & 74.9 & \textbf{77.5} & +2.4 & 1.02$\times$\\
            R(2+1)D-50 16$\times$2~\cite{tran2017closer} & 75.7 &  \textbf{78.0} & +2.3 & 1.01$\times$\\
            \midrule
            SlowFast-50 4$\times$16~\cite{feichtenhofer2019slowfast} & 75.6 & \textbf{77.8} & +2.2 & 1.02$\times$\\
            SlowFast-50 8$\times$8~\cite{feichtenhofer2019slowfast} & 76.8 &  \textbf{78.9} & +2.1 & 1.03$\times$\\
            \bottomrule
            \vspace{3pt}
        \end{tabular*}
        \vspace{-0.2in}
        \captionof{table}{Generalization to \textbf{different backbone architectures}. We report the video-level accuracy on Kinetics-400 for both standard clip-level training (``Baseline") and video-level training with \cm (``Ours"). }
        \label{tab:backbones}
    \end{minipage}
    \vspace{-0.1in}
\end{figure*}

\subsection{Coping with the GPU Memory Constraint}\label{sec:gpu}
A challenge posed by video-level learning is the need to jointly optimize over multiple clips under a fixed and tight GPU memory budget. In this section, we discuss two simple implementations that allow end-to-end training of video-level dependencies under this constraint. 

\vspace{-0.17in}
\paragraph{Batch reduction.} Let $B$ be the size of the mini-batch of videos used for traditional clip-level training. Our approach can be implemented under the same GPU memory budget  by reducing the batch size by a factor of $N$: $\hat{B} = round(B/N)$. This allows us to load into the memory $N$ clips for each of the $B/N$ different videos. In order to improve the clip diversity for updating the batch-norm~\cite{IoffeS15} parameters within a mini-batch, we propose to calculate batch-norm statistics using only the clips from different videos. Although this implementation cannot handle arbitrarily large $N$, it is simple, efficient and we found it applicable to most settings in practice. For example, a typical choice of batch size for training clip-based models is a 8-GPU machine with $B=64$; our approach can be implemented under this memory setup by using in each iteration $\hat{B}=16$ different videos, and by sampling from each of them $N=4$ clips. 

\vspace{-0.17in}
\paragraph{Multi-iteration.} Instead of directly loading $N$ clips into one mini-batch, we can also unroll the training of a video into $N$ iterations. Each iteration uses one of the $N$ clips. 
This implementation is memory-friendly and consumes the same amount of memory as the standard single-clip training framework. It allows us to process arbitrarily long videos with arbitrarily large $N$. When incorporating the collaborative memory, we simply perform a two-scan process: the first scan generates the clip-based features to build the global memory $\mathcal{M}$ and the second scan generates the classification output of each clip conditioned on $\mathcal{M}$.


%% file: setup.tex
\section{Experiments}\label{sec:setup}
\vspace{-0.05in}
To demonstrate the advantages of our end-to-end video-level learning framework, we conduct extensive experiments on four action recognition benchmarks with different backbone architectures. 
We implement our models and conduct the experiments using the  PySlowFast codebase~\cite{fan2020pyslowfast}. 

\subsection{Experimental Setup}
\paragraph{Datasets.}
Kinetics~\cite{kinetics} (K400 \& K700) is one of the most popular datasets for large-scale video classification. Charades~\cite{sigurdsson2016hollywood} is a multi-label dataset with long-range activities. Something-Something-V1~\cite{goyal2017something} is a dataset requiring good use of temporal relationships for accurate recognition. Following the standard protocol, we use the training set for training and report top-1 accuracy on the validation set.
\vspace{-0.17in}
\paragraph{Backbones.} We evaluate our framework using different backbone architectures including I3D~\cite{wang2017non}, R(2+1)D~\cite{tran2017closer,wang2020video}, Slow-only~\cite{feichtenhofer2019slowfast} and SlowFast~\cite{feichtenhofer2019slowfast}, optionally augmented with non-local blocks (NL)~\cite{wang2017non}. 
We attach the proposed collaborative memory to the last convolutional layer of these backbones for joint training.

\vspace{-0.17in}
\paragraph{Training.} We first train the backbones by themselves following their original schedules~\cite{feichtenhofer2019slowfast,wang2020video}, then re-train the backbones in conjunction with our collaborative memory for video-level learning. 
When training on Kinetics, we use synchronous SGD with a cosine learning schedule~\cite{loshchilov2016sgdr}. 
For Charades and Something-Something-V1, we follow the recipe from PySlowFast~\cite{fan2020pyslowfast} and initialize the network weights from the models pre-trained on Kinetics, since these two datasets are relatively small.
For the video-level training,  we employ the batch reduction strategy to handle the GPU memory constraint by default and apply the linear scaling rule~\cite{goyal2017accurate} to adjust the training schedule accordingly. 





\vspace{-0.17in}
\paragraph{Inference.} Following~\cite{wang2017non,feichtenhofer2019slowfast}, we uniformly samples $3\times 10$ crops from each video for testing (\ie, $3$ spatial crops and $10$ temporal crops).
The global memory $\mathcal{M}$ is aggregated from 10 spatially centered crops and shared for the inference of all 30 crops.
We employ the multi-iteration approach from Section~\ref{sec:gpu} during inference to overcome the GPU memory constraint.
The softmax scores of all $30$ clips are averaged for the final video-level prediction.


%% file: exp.tex

\subsection{Evaluating Collaborative Memory}


For all the experiments in this section we use the associative version of the collaborative memory with feature gating since, as demonstrated in ablation studies (Section~\ref{sec:ablation}), this design provides the best results.

\begin{table*}[t!]
    \begin{subtable}[b]{0.3\linewidth}
        \centering
        \footnotesize
        \renewcommand{\arraystretch}{1.2}
        \setlength{\tabcolsep}{0pt} 
        \begin{tabular*}{\linewidth}{@{\extracolsep{\fill}\;}cccc@{\extracolsep{\fill}\;}}
            \toprule
            \textbf{Multi-clip} & \textbf{Memory} & \textbf{End-to-end} &  \textbf{Top-1} \\
            \midrule
             &\checkmark &\checkmark & 74.5 \\
             \checkmark & & \checkmark & 75.5  \\
            \checkmark & \checkmark &  & 75.9   \\ 
            \midrule
            \checkmark & \checkmark & \checkmark & \textbf{76.8} \\
            \bottomrule
        \end{tabular*}
        \caption{Evaluating \textbf{different components} of our video-level learning framework.}
        \label{tab:ablation_components}
        \vspace{0.05in}
    \end{subtable}
    \hfill
    \begin{subtable}[b]{0.38\linewidth}
        \centering
        \footnotesize
        \renewcommand{\arraystretch}{1.2}
        \setlength{\tabcolsep}{0pt} 
        \begin{tabular*}{\linewidth}{@{\extracolsep{\fill}\;}lccc@{\extracolsep{\fill}\;}}
            \toprule
            \multicolumn{1}{c}{\textbf{Setting}} & \textbf{Associative} & \textbf{Gating}  &  \textbf{Top-1}\\
            \midrule
            Multi-clip (w/o memory) &  &  & 75.5  \\
            \midrule
            CM (avgpool) &  &\checkmark  &  75.8 \\
            CM (residual) & \checkmark & & 76.0 \\
            CM (default) & \checkmark & \checkmark & \textbf{76.8}  \\
            \bottomrule
        \end{tabular*}
        \caption{Comparing \textbf{different designs} of our collaborative memory mechanism.}
        \label{tab:ablation_memory}
        \vspace{0.05in}
    \end{subtable}
    \hfill
    \begin{subtable}[b]{0.22\linewidth}
        \centering
         \footnotesize
         \renewcommand{\arraystretch}{1.2}
         \setlength{\tabcolsep}{0pt} 
         \begin{tabular*}{\linewidth}{@{\extracolsep{\fill}\;}lcc@{\extracolsep{\fill}\;}}
            \toprule
            \multicolumn{1}{c}{\textbf{Setting}} & \textbf{\#Param.} & \textbf{Top-1} \\
             \midrule
             $\alpha=1$  & 49.2 M & 76.8 \\
             $\alpha=2$  & 40.9 M & 76.8 \\
             $\alpha=4$  & 36.7 M & \textbf{76.8} \\
             $\alpha=8$  & 34.6 M & 76.4 \\
             \bottomrule
         \end{tabular*}
         \caption{Varying \textbf{channel reduction ratio} $\alpha=d/d^{'}$.}
         \label{tab:ablation_alpha}
         \vspace{0.05in}
    \end{subtable}
    
    \begin{subtable}[b]{0.27\linewidth}
        \centering
        \footnotesize
        \renewcommand{\arraystretch}{1.2}
        \setlength{\tabcolsep}{0pt} 
        \begin{tabular*}{\linewidth}{@{\extracolsep{\fill}\;}lcc@{\extracolsep{\fill}\;}}
        \toprule
        \multicolumn{1}{c}{\textbf{Model}} & {\footnotesize \bf Stage-wise} & \textbf{Top-1} \\
         \midrule
         \multirow{2}{*}{Slow-only} &  & 76.1 \\
          & \checkmark & \textbf{76.8}   \\
         \midrule
         \multirow{2}{*}{R(2+1)D} &  & 77.7  \\
         & \checkmark  & \textbf{78.0} \\
         \bottomrule
        \end{tabular*}
        \caption{\textbf{Stage-wise training} \vs training everything from scratch.}
        \vspace{-0.05in}
        \label{tab:ablation_stage}
    \end{subtable}
    \hfill    
    \begin{subtable}[b]{0.35\linewidth}
        \centering
        \footnotesize
        \renewcommand{\arraystretch}{1.2}
        \setlength{\tabcolsep}{0pt} 
        \begin{tabular*}{\linewidth}{@{\extracolsep{\fill}\;}lccc@{\extracolsep{\fill}\;}}
            \toprule
            \multicolumn{1}{c}{\textbf{Model}} & {\footnotesize \bf Batch reduction} & {\footnotesize \bf Multi-iteration} & \textbf{Top-1} \\
            \midrule
            \multirow{2}{*}{Slow-only} & &\checkmark & 76.6 \\
                & \checkmark & & \textbf{76.8}   \\
            \midrule
            \multirow{2}{*}{R(2+1)D} & &\checkmark & 77.9 \\
                & \checkmark & & \textbf{78.0}   \\
            \bottomrule
        \end{tabular*}
        \caption{Comparing different ways of training CM: \textbf{batch reduction} \vs multi-iteration.}
        \vspace{-0.05in}
        \label{tab:ablation_training}
    \end{subtable}
    \hfill
    \begin{subtable}[b]{0.3\linewidth}
        \centering
        \footnotesize
        \renewcommand{\arraystretch}{1.2}
        \setlength{\tabcolsep}{0pt} 
        \begin{tabular*}{\linewidth}{@{\extracolsep{\fill}\;}lccccc@{\extracolsep{\fill}\;}}
            \toprule
            \multicolumn{1}{c}{\multirow{2}{*}{\bf Model}} & \multicolumn{4}{c@{}}{Temporal stride} & \multicolumn{1}{c}{\multirow{2}{*}{\bf CM}}\\
            \cmidrule{2-5}
             & \textbf{2} & \textbf{4}  & \textbf{8} & \textbf{16} & \\
            \midrule 
            Slow-only & 73.2 & 74.3 & 74.4 & 74.4 & \textbf{76.8}\\
            \midrule
            R(2+1)D & 75.7 & 76.4 & 75.0 & 72.2 & \textbf{78.0} \\
            \bottomrule
        \end{tabular*}
        \caption{Comparing CM with backbones using clips with \textbf{large temporal strides}.}
        \vspace{-0.05in}
        \label{tab:ablation_stride}
    \end{subtable}
    \caption{Ablation experiments on Kinetics-400. Top-1 video-level accuracy (\%) is reported. Unless otherwise stated, 
    we use Slow-only~\cite{feichtenhofer2019slowfast} with 50 layers and the input clip length is $8\times 8$. R(2+1)D is also 50 layers with a clip length of $16\times 2$.}
    \vspace{-0.1in}
\end{table*}

\vspace{-0.17in}
\paragraph{Effectiveness of video-level learning.} We begin by presenting an experimental comparison between our proposed video-level learning and the standard clip-level training applied to the same architecture. For this evaluation we use the Slow-only  backbone with 50 layers, which can be considered as a 3D ResNet~\cite{KaimingHe16}. 
In order to investigate the impact of temporal coverage on video-level learning, we train models using different numbers of sampled clips per video: $N\in\{1, 3, 5, 7, 9\}$. $N=1$ corresponds to the conventional clip-level training, as we only sample one clip per video. In such case the collaborative memory (\cm) is limited to perform ``self-attention" within the single clip. For $N>1$, \cm captures video-level information across the $N$ clips. 



Figure~\ref{fig:main} shows the {\em video}-level accuracy achieved by the different models, all using the same testing setup of $3\times 10$ crops per video. 
Note that under this setting all models ``look'' at the same number of clips for each video in testing. As shown in Figure~\ref{fig:main}, our \cm framework significantly improves the video-level accuracy. 
For example, when the clip length is 8$\times$8 (8 frames with a temporal stride of 8), using \cm with $N=9$ yields a remarkable $2.6\%$ improvement compared with training using a single clip 
($74.5\%$ \vs $77.1\%$). When the clip has a shorter length (\ie, 8$\times$2), our method gives an even larger gain, $3.2\%$ ($73.2\%$ \vs $76.4\%$). As expected, the improvement from our method increases with the number of sampled clips $N$. The performance saturates when $N\geq 7$. To keep the training time more manageable, we use $N=5$ by default.



Figure~\ref{fig:clip} shows the clip-level accuracy at different temporal locations of a video, according to their temporal order. 
When $N=1$, the clips from the middle of the video have much higher accuracy than the clips from the beginning or the end of the video, as the middle clips tend to include more relevant information. 
\cm significantly improves the clip-level accuracy by augmenting each clip with video-level context information (\ie, $N\geq3$), 
especially for clips near the boundary of the video. This is a clear indication that our memory mechanism is capable of capturing video-level dependencies and sharing them effectively with the clips within the video to boost the recognition accuracy.


\vspace{-0.1in}
\paragraph{Generalization to different backbones.}
As we do not make any assumption about the backbone,  our video-level end-to-end learning framework can be seamlessly integrated with different architectures and input configurations (\eg, clip length, sampling stride, \etc).
As shown in Table~\ref{tab:backbones}, \cm produces consistent video-level accuracy gains of over 2\% on top of state-of-the-art video models, including I3D with non-local blocks~\cite{wang2017non}, the improved R(2+1)D network~\cite{tran2017closer,wang2020video} and the recent SlowFast network~\cite{feichtenhofer2019slowfast}. Note that we achieve these improvements with only negligible additional inference cost, about $1$-$3\%$ more FLOPs compared to the backbone themselves.

\subsection{Ablation Studies}\label{sec:ablation}


\noindent\textbf{Assessing the components in our framework.}
Unlike most prior work on video-level modeling~\cite{hussein2019timeception,wu2019long}, our framework is end-to-end trainable. To show the benefits of end-to-end learning in improving the backbone, we conduct an ablation that freezes the parameters of the backbone 
and only updates the parameters from the collaborative memory and the FC layers for classification. 
As shown in Table~\ref{tab:ablation_components}, end-to-end learning improves the performance by 1.1\% compared with learning video-level aggregation on top of the frozen backbones (76.8\% \vs 75.9\%).

Table~\ref{tab:ablation_components} also shows the result of video-level learning without using \cm. This is done by optimizing multiple clips per video but without sharing any information across the clips. Interestingly, this simple setup also delivers a good improvement over the single-clip learning baseline (75.5\% vs. 74.5\%). 
The gain comes from the joint optimization over multiple clips of a video, which again confirms the importance of video-level learning for classification. Our CM framework achieves the best performance with all the components enabled. 

\vspace{-0.1in}
\paragraph{Collaborative memory design.}
Our default design uses the associative memory for memory interactions and a feature gating operation for context infusion. In Table~\ref{tab:ablation_memory}, we explore other design choices by replacing the associative memory with average pooling or substituting the feature gating with a residual connection.

\begin{figure}[t]
    \centering
    \includegraphics[width=0.85\linewidth]{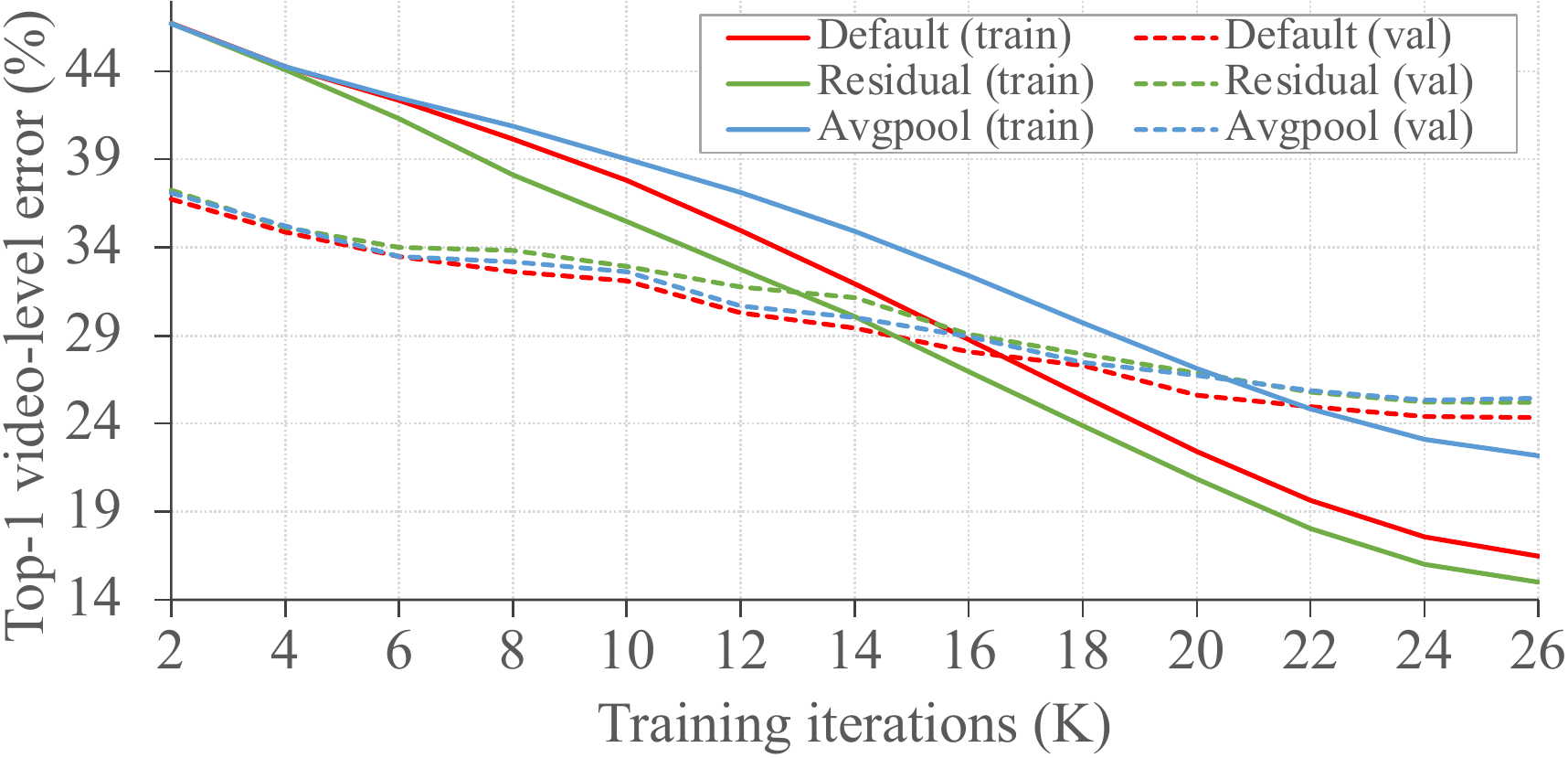}
    \vspace{-0.05in}
    \caption{Video-level training/validation errors on Kinetics-400 for different designs of the collaborative memory.}
    \label{fig:curves}
    \vspace{-0.13in}
\end{figure}

We observe that all these variants offer improvement over the na\"{\i}ve video-level learning setup without the memory. However, the gain provided by average pooling is relatively small, which is not surprising given that there is no the inter-clip interaction.
While we also witness a performance drop by removing the feature gating operation, the reason behind it is different. As shown by the training/validation error curves in Figure~\ref{fig:curves}, the model without feature gating achieves lower training error but higher validation error. This suggests that the model degenerates during training due to over-fitting to the video-level context.

In Table~\ref{tab:ablation_alpha} we ablate the number of channels used in \cm ($d'$ in Eq.~\ref{eq:push},~\ref{eq:pop}), which can be controlled by the channel reduction ratio $\alpha=d / d'$. We can see that the results remain unchanged as long as the reduction ratio is reasonable ($\alpha \leq 4)$. We use $\alpha=4$ as the default value in our experiments since it introduces fewer extra parameters.

\vspace{-0.1in}
\paragraph{Training strategies.} \label{sec:training}
Recall that we adopt a stage-wise training strategy: the backbones are first trained using standard clip-level training recipes and then re-trained in conjunction with \cm for video-level learning. In Table~\ref{tab:ablation_stage}, we compare this strategy with training everything from scratch (equivalent training iterations are used for both strategies). Experiments on two different backbones show that training everything from scratch yields slightly worse results.
We hypothesize that stage-wise training allows the optimization in the second stage to focus on effective long-range modeling thanks to the well-initialized backbone. We note that non-local networks are also trained in a stage-wise way. 

We also compare the two methods to cope with the GPU memory constraint (Section~\ref{sec:gpu}). As shown in Table~\ref{tab:ablation_training}, the accuracy of the two methods is almost the same and the difference is within the margin of randomness, which makes sense as the two methods are technically identical. 

\vspace{-0.1in}
\paragraph{Limitations of temporal striding.}
One simple way to increase the temporal coverage of a video model 
is to use larger temporal strides when sampling the frames of the input clips. 
We compare our video-level learning framework with this strategy in Table~\ref{tab:ablation_stride}. Note that we keep the temporal strides of \cm the same as the original backbones, \ie, $8$ frames for Slow-only and $2$ frames for R(2+1)D. 

We can see that increasing the temporal coverage through striding yields only a modest gain in accuracy. Notably, using a very large stride even hurts performance for some architectures like R(2+1)D. 
In contrast, our approach can learn long-range dependencies and improves the performances of short clip-based backbones by large margins.

\subsection{Comparison with the State of the Art}
Previous experimental results are from Kinetics-400. To demonstrate that our method can generalize to different datasets, we further evaluate our method on Kinetics-700~\cite{carreira2019short}, Charades~\cite{sigurdsson2016hollywood} and Something-Something-V1~\cite{goyal2017something}.
Among them, Charades has longer-range activities (30 seconds on average), 
whereas Something-Something-V1 includes mostly human-object interactions. 
We compare the results with the state of the art in Table~\ref{tab:sota_k400},~\ref{tab:sota_k700},~\ref{tab:sota_charades} and~\ref{tab:sota_sthv1}. Our proposed \cm framework yields consistent gains of over $2\%$ for different variants of SlowFast on all datasets. These improvements are very significant given that SlowFast is among the best video backbones.

\begin{table}[t]
    \centering
    \footnotesize
    \renewcommand{\arraystretch}{1.2}
    \setlength{\tabcolsep}{0pt} 
    \begin{tabular*}{\linewidth}{@{\extracolsep{\fill}\,}lcccc@{}}
        \toprule
		\multicolumn{1}{c}{\multirow{2}{*}{\bf Methods}}  & \multirow{2}{*}{\bf Pretrain} & {\bf Only} & {\bf GFLOPs} & \multirow{2}{*}{\bf Top-1}  \\ [-3pt]
			 & &  {\bf RGB} & {\bf $\times$ crops} &  \\
		\midrule
			I3D~\cite{I3D} & ImageNet & \redxmark & 216$\times$N/A & 75.7  \\
			S3D-G~\cite{xie2017rethinking}  & ImageNet & \redxmark & 142.8$\times$N/A  & 77.2  \\ 
			LGD-3D-101~\cite{qiu2019learning} & ImageNet & \redxmark & N/A& 81.2 \\ 
			I3D-101+NL~\cite{wang2017non} & ImageNet & \greencmark & 359$\times$30 & 77.7 \\
			ip-CSN-152~\cite{tran2019video} & 	 Sports1M & \greencmark & 109$\times$30 & 79.2 \\
			CorrNet-101 & Sports1M & \greencmark & 224$\times$30 & 81.0   \\
			\midrule
			MARS+RGB~\cite{crasto2019mars} & none & \greencmark & N/A &  74.8   \\
			DynamoNet~\cite{diba2019dynamonet}  & 	 none & \greencmark & N/A & 77.9 \\
			CorrNet-101~\cite{wang2020video} & none & \greencmark & 224$\times$30 & 79.2 \\
            SlowFast-101 8$\times$8~\cite{feichtenhofer2019slowfast} & none & \greencmark  & 106$\times$30 & 77.9 \\
			SlowFast-101 16$\times$8~\cite{feichtenhofer2019slowfast} & none & \greencmark & 213$\times$30 & 78.9 \\
			SlowFast-101+NL 16$\times$8~\cite{feichtenhofer2019slowfast} & none & \greencmark  & 234$\times$30 & 79.8 \\
		\hline\hline
        {\bf Ours} (R(2+1)D-101 32$\times$2) & none & \greencmark & 243$\times$30 &80.5 \\
        {\bf Ours} (SlowFast-101 8$\times$8) & none & \greencmark & 128$\times$30 &80.0 \\
        {\bf Ours} (SlowFast-101+NL 8$\times$8) & none & \greencmark & 137$\times$30 &\textbf{81.4} \\
        \bottomrule
    \end{tabular*}
    \vspace{-0.07in}
    \caption{Comparison with the state-of-the-art on Kinetics-400.}
    \label{tab:sota_k400}
    \vspace{-0.05in}
\end{table}

\begin{table}[t]
    \centering
    \footnotesize
    \renewcommand{\arraystretch}{1.2}
    \setlength{\tabcolsep}{0pt} 
    \begin{tabular*}{\linewidth}{@{\extracolsep{\fill}\,}lccc@{}}
        \toprule
		\multicolumn{1}{c}{\multirow{2}{*}{\bf Methods}} & \multirow{2}{*}{\bf Pretrain} & {\bf GFLOPs} & \multirow{2}{*}{\bf Top-1} \\ [-3pt] 
		 & &  {\bf $\times$ crops} & \\
		\midrule
        SlowFast-101+NL 8$\times$8~\cite{feichtenhofer2019slowfast} & K600  & 115$\times$30 & 70.6 \\
        SlowFast-101+NL 16$\times$8~\cite{feichtenhofer2019slowfast} & K600 & 234$\times$30 & 71.0 \\
        \midrule        
        SlowFast-50 4$\times$16$^*$ & K600 & 36$\times$30 & 66.1 \\
        SlowFast-101 8$\times$8$^*$ & K600& 126$\times$30 & 69.2 \\
        SlowFast-101+NL 8$\times$8$^*$ & K600& 135$\times$30 &  70.2 \\
        \hline\hline
        {\bf Ours} (SlowFast-50 4$\times$16) & K600& 37$\times$30 &68.3  \\
        {\bf Ours} (SlowFast-101 8$\times$8) & K600& 128$\times$30 & 70.9 \\
        {\bf Ours} (SlowFast-101+NL 8$\times$8) & K600& 137$\times$30 &\textbf{72.4} \\
        \bottomrule
    \end{tabular*}
    \vspace{-0.05in}
    \caption{Comparison with the state-of-the-art on Kinetics-700. $^*$ indicates results reproduced by us.}
    \label{tab:sota_k700}
    \vspace{-0.15in}
\end{table}

On Kinetics-400 and Kinetics-700, our method establishes a new state of the art, as shown in Table~\ref{tab:sota_k400} and~\ref{tab:sota_k700}. Notably, we achieve these results without pre-training on other datasets or using optical flow.  Similarly, our method outperforms the state of the art on both Charades (in Table~\ref{tab:sota_charades}) and Something-Something-V1 (in Table~\ref{tab:sota_sthv1}). On Charades, our CM framework yields more than 3\% gains (\eg, 44.6\% vs 41.3\%).
This demonstrates that \cm performs even better on datasets that have longer videos and require longer-term temporal learning. Note that our method significantly outperforms other recent work on long-range temporal modeling (\eg, {Timeception~\cite{hussein2019timeception}, LFB~\cite{wu2019long}}).

\begin{table}[t]
    \centering
    \footnotesize
    \renewcommand{\arraystretch}{1.2}
    \setlength{\tabcolsep}{0pt} 
    \begin{tabular*}{\linewidth}{@{\extracolsep{\fill}\,}lccc@{}}
        \toprule
		\multicolumn{1}{c}{\multirow{2}{*}{\bf Methods}} & \multirow{2}{*}{\bf Pretrain} & {\bf GFLOPs} & \multirow{2}{*}{\bf Top-1} \\ [-3pt]
		 & &  {\bf $\times$ crops} & \\
		\midrule
		TRN~\cite{zhou2018temporal} & ImageNet & N/A & 25.2 \\
		I3D-101+NL~\cite{wang2017non} & ImageNet+K400  & 544 $\times$ 30 & 37.5\\
		STRG~\cite{wang2018videos} & ImageNet+K400 &  630 $\times$ 30 &39.7 \\
		Timeception~\cite{hussein2019timeception} & K400 & N/A & 41.1 \\
		LFB (I3D-101+NL)~\cite{wu2019long} & K400 & N/A & 42.5 \\
        SlowFast-101+NL~\cite{feichtenhofer2019slowfast} & K400 & 234$\times$30 & 42.5 \\
        AVSlowFast-101+NL~\cite{xiao2020avslowfast} & K400 & 278$\times$30 & 43.7 \\
        \midrule        
        SlowFast-50 16$\times$8$^*$ & K400 & 131$\times$30 & 39.4 \\
        SlowFast-101+NL 16$\times$8$^*$ & K400& 273$\times$30 &  41.3 \\
        \hline\hline
        {\bf Ours} (SlowFast-50 16$\times$8) & K400& 135$\times$30 & 42.9  \\
        {\bf Ours} (SlowFast-101+NL 16$\times$8) & K400& 277$\times$30 &\textbf{44.6} \\
        \bottomrule
    \end{tabular*}
    \vspace{-0.05in}
    \caption{Comparison with the state-of-the-art on Charades. 
    $^*$ indicates results reproduced by us.}
    \vspace{-0.05in}
    \label{tab:sota_charades}
\end{table}

\begin{table}[t]
    \centering
    \footnotesize
    \renewcommand{\arraystretch}{1.2}
    \setlength{\tabcolsep}{0pt} 
    \begin{tabular*}{\linewidth}{@{\extracolsep{\fill}\,}lccc@{}}
        \toprule
		\multicolumn{1}{c}{\bf Methods}  & {\bf Pretrain} & {\bf Only RGB} & {\bf Top-1}  \\ 
		\midrule
            S3D-G~\cite{xie2017rethinking}  & ImageNet & \redxmark &   48.2   \\ 
			ECO~\cite{zolfaghari2018eco} & none & \redxmark &  49.5  \\ 
            Two-stream TSM~\cite{lin2019tsm} & ImageNet & \redxmark & 52.6  \\
			MARS+RGB+Flow~\cite{crasto2019mars} & K400 & \redxmark & 53.0 \\
			NL I3D-50+GCN~\cite{wang2018videos} & ImageNet & \greencmark &  46.1  \\
			GST-50~\cite{luo2019grouped} & ImageNet & \greencmark &  48.6   \\
			MSNet~\cite{kwon2020motionsqueeze} & ImageNet & \greencmark & 52.1  \\
			CorrNet-101~\cite{wang2020video} & Sports1M & \greencmark &  53.3  \\
			\midrule
			SlowFast-50 8$\times$8* & K400 & \greencmark & 50.1  \\
            SlowFast-101+NL 8$\times$8* & K400 & \greencmark  & 51.2   \\
		\hline\hline
        {\bf Ours} (SlowFast-50 8$\times$8) & K400 & \greencmark & 52.3   \\
        {\bf Ours} (SlowFast-101+NL 8$\times$8) & K400 & \greencmark  &\textbf{53.7}  \\
        \bottomrule
    \end{tabular*}
    \vspace{-0.05in}
    \caption{Comparison with the state-of-the-art on Something\\-Something-V1. $^*$ indicates results reproduced by us.}
    \label{tab:sota_sthv1}
    \vspace{-0.15in}
\end{table}

\subsection{Collaborative Memory for Action Detection}

In this section, we show that the benefits of our framework also extend to the task of action detection. We evaluate our method on AVA~\cite{gu2018ava}, which includes 211k training and 57k validation video segments. AVA v2.2 provides more consistent annotations than v2.1 on the same data. We report mean average precision (mAP) over 60 frequent classes on the validation set following the standard protocol.

\vspace{-0.1in}
\paragraph{Adaptation to action detection.} 
Adapting our approach to action detection is straightforward. Instead of randomly sampling multiple clips from the whole video,  we sample clips within a certain temporal window $t_n \in [t-w, t+w]$ to detect action at time $t$, where $t_n$ indicates the center frame of the $n$th sampled clip and $2w+1$ is the window size. As AVA includes sparse annotations at one frame per second, we simply sample the clips with a one-second stride such that the sampled clips are centered at frames with annotations. In this way, the temporal window size increases accordingly as we use a larger number of clips $N$ during training. After that, we jointly optimize these sampled clips with their own annotations. The collaborative memory is used to share long-range context information among sampled clips. 
We use $N=9$ in our experiments and follow the schedule in AIA~\cite{tang2020asynchronous} for model training. 


\begin{table}[t]
    \begin{subtable}[b]{0.47\linewidth}
        \centering
        \footnotesize
        \renewcommand{\arraystretch}{1.2}
        \setlength{\tabcolsep}{0pt} 
        \begin{tabular*}{\linewidth}{@{}lcc}
            \toprule
            \multicolumn{1}{c}{\bf Methods}  & {\bf Pretrain\,} & {\bf mAP}  \\
             \midrule
            ACRN~\cite{Sun2018ACRN} & K400  & 17.4$^{\dagger}$   \\
            AVSF-50 4$\PLH$16~\cite{xiao2020avslowfast} & K400  & 27.8$^{\dagger}$  \\
            AT (I3D)~\cite{Rohit2019video} & K400 & 25.0 \\
            LFB(R50+NL)~\cite{wu2019long} & K400 &  25.8 \\
            \midrule
            R50+NL$^*$~\cite{wu2019long} & K400 & 23.6 \\
            SF-50 4$\PLH$16$^*$~\cite{feichtenhofer2019slowfast} & K400 & 23.6  \\
            \hline \hline
            {\bf Ours} (R50+NL) & K400  & 26.3  \\
            {\bf Ours} (SF-50 4$\PLH$16) & K400  & 25.8 \\
            \bottomrule
        \end{tabular*}
        \caption{}
        \vspace{-0.1in}
    \end{subtable}
    \hfill
    \begin{subtable}[b]{0.51\linewidth}
        \centering
        \footnotesize
        \renewcommand{\arraystretch}{1.2}
        \setlength{\tabcolsep}{0pt} 
        \begin{tabular*}{\linewidth}{@{}lcc}
            \toprule
            \multicolumn{1}{c}{\bf Methods}  & {\bf Pretrain\,} & {\bf mAP}  \\
             \midrule
            AVSF-101 8$\PLH$8~\cite{xiao2020avslowfast} & K400  & 28.6$^{\dagger}$  \\
            AIA(SF-50 4$\PLH$16)~\cite{tang2020asynchronous} & K700 &  29.8$^{\dagger}$  \\
            AIA(SF-101 8$\PLH$8)~\cite{tang2020asynchronous} & K700 &  32.3$^{\dagger}$  \\
            SF-101+NL 8$\PLH$8~\cite{feichtenhofer2019slowfast} & K600 &  29.0 \\
            \midrule
            SF-50 4x16$^*$~\cite{feichtenhofer2019slowfast} & K700 &  26.9 \\
            SF-101 8x8$^*$~\cite{feichtenhofer2019slowfast} & K700 & 29.0 \\
            \hline \hline
            \textbf{Ours} (SF-50 4$\PLH$16) & K700 & 29.8 \\
            \textbf{Ours} (SF-101 8$\PLH$8) & K700 & 31.6 \\
            \bottomrule
        \end{tabular*}
        \caption{}
        \vspace{-0.1in}
    \end{subtable}
    \caption{Comparison with SOTA on (a) AVA v2.1 and (b) v2.2. $^{\dagger}$ indicates results with extra information other than RGB frames, such as optical flow, audio and objection detection predictions. $^*$ indicates results reproduced by us.}
    \label{tab:ava_sota}
    \vspace{-0.15in}
\end{table}

\vspace{-0.1in}
\paragraph{Quantitative results.} 
We compare \cm with the state of the art on AVA in Table~\ref{tab:ava_sota}. Although the \cm framework is not specifically designed for action detection, it achieves results comparable with the state of the art. 
In particular, \cm yields a consistent improvement of more than $2\%$ for different backbone networks (\eg, 2.9\% gain for SlowFast-50 4$\times$16 backbone on AVA v2.2).
This demonstrate that we can effectively extend our method to the detection task and achieve significant improvements as well.
Note that our method also outperforms LFB~\cite{wu2019long} when using the same backbone (\ie, R50-I3D+NL) on AVA v2.1 (26.3\% \vs 25.8\%).




%% file: conclusion.tex
\section{Conclusions}\label{sec:conclusion}

We have presented an end-to-end learning framework that optimizes classification models using video-level information. 
Our approach hinges on a novel collaborative memory mechanism that accumulates contextual information from multiple clips sampled from the video and then shares back this video-level context to enhance the individual clip representations. Long-range temporal dependencies beyond short clips are captured through the interactions between the local clips and the global memory.
Extensive experiments on both action recognition and detection benchmarks show that our framework significantly improves the accuracy of video models at a negligible computational overhead.

%% file: appendix.tex
\appendix
\section*{Appendix}
We provide additional results on AVA in Section~\ref{sec:supp_ava} and more implementation details on the each dataset in Section~\ref{sec:supp_details}.

\section{Additional results on AVA} \label{sec:supp_ava}
In addition to the comparison with the state-of-the-art, we also investigate how the performance changes \wrt the number of clips $N$ used in \cm on AVA v2.2. As shown in Table~\ref{tab:ava_temporal}, increasing $N$ does improve the mAP on AVA, similar to the observation from Kinetics (see Figure~\ref{fig:main}). When $N$ is 9, \cm is $2.9\%$ better than using a single clip (\ie, $N=1$). This again demonstrate that we can effectively extend \cm to different tasks (such as action localization) and achieve significant improvements.

\begin{table}[h]
     \centering
     \footnotesize
     \renewcommand{\arraystretch}{1.2}
     \setlength{\tabcolsep}{0pt} 
     \begin{tabular*}{\linewidth}{@{\extracolsep{\fill}\quad}lccccc}
         \toprule
         \textbf{N} & \textbf{1} & \textbf{3}  & \textbf{5} & \textbf{7} & \textbf{9}\\
         \midrule
         mAP (\%) & 26.9  & 27.4 & 28.3 & 29.3 & \textbf{29.8}  \\
         \bottomrule
     \end{tabular*}
     \caption{Accuracy on AVA v2.2 when using different number of clips $N$ in \cm for both training and testing.}
     \label{tab:ava_temporal}
\end{table}

\section{Implementation Details} \label{sec:supp_details}

\subsection{Kinetics}
\paragraph{Data setup.} Kinetics-400~\cite{kinetics} consists of about 300K YouTube videos covering 400 categories. Kinetics-700~\cite{carreira2019short} is the latest version with  650k videos and 700 categories. We densely sample the input clips from the full video with temporal jittering, and then perform data augmentation such as scale-jittering, horizontal flipping and random cropping. We use a scale jittering range of $[256, 320]$ for models with 50 layers and $[256, 340]$ for those with 101 layers following~\cite{feichtenhofer2019slowfast}, and randomly crop $224\times224$ pixels for training. The mini-batch size is 64 clips per node (8 GPUs).

\vspace{-0.1in}
\paragraph{Backbone training.} We use backbones including I3D~\cite{wang2017non}, R(2+1)D~\cite{tran2017closer,wang2020video}, Slow-only~\cite{feichtenhofer2019slowfast} and SlowFast~\cite{feichtenhofer2019slowfast}, optionally augmented with non-local blocks (NL)~\cite{wang2017non}. Different backbones are trained following their original schedules and we summarize them here for clarity. We use a momentum of 0.9 and a weight decay of $10^{-4}$ for all models. Dropout~\cite{hinton2012improving} of 0.5 is used before the final classifier layer.

The training of I3D, Slow-only and SlowFast follows the recipe from~\cite{feichtenhofer2019slowfast}.
For Slow-only and SlowFast, we use synchronous SGD with a cosine learning rate schedule~\cite{loshchilov2016sgdr}. The model is trained for 196 epochs with the first 34 epochs for linear warm-up~\cite{goyal2017accurate}. The initial learning rate is set as 0.1 per node (8 GPUs). For the models augmented with NL, we initialize them with the counterparts that are trained without NL for easier optimization~\cite{feichtenhofer2019slowfast}. Similarly, only 3 NL blocks are used on the Slow features of res$_4$.
For I3D-50+NL, we initialize the model with ImageNet~\cite{deng2009imagenet} pretrained weights and train the model for 118 epochs using a step-wise learning rate schedule. The initialize learning rate is set to 0.01 per node and is reduced by a factor of 10 at 44, 88 epochs~\cite{fan2020pyslowfast}.

For R(2+1)D, we follow the recipe in~\cite{wang2020video}. Specifically, we train the model for 250 epochs using the cosine learning rate schedule, with the first 50 epochs for linear warm-up. The initial learning rate is set as 0.4 per node.

For Kinetics-700, we initialize the model with the weights pretrained on Kinetics-600 provided by PySlowFast~\cite{fan2020pyslowfast} and train the model for 240 epochs using the cosine learning rate schedule with the first 34 epochs for linear warm-up. The initial learning rate is set to 0.05 per node.

\vspace{-0.1in}
\paragraph{Video-level training with \cm.} We re-train the initialized backbones with \cm for video-level learning as described in Section~\ref{sec:training}. Note that the \textit{same} training schedule is used for all the backbones during re-training. The linear scaling rule~\cite{goyal2017accurate} is applied to adjust the training schedule accordingly for different numbers of clips $N$ sampled per video. Take $N=5$ as an example, we first adjust the mini-batch size per GPU to $\hat{B}=2$ following the batch reduction strategy in Section~\ref{sec:gpu}. The model is then trained for 15 epochs using the cosine learning rate schedule, with the initial learning rate of 0.0125 per node and 2.25 epochs for linear warm-up. 

\subsection{Charades}
Charades~\cite{sigurdsson2016hollywood} consists of about 9.8k training and 1.8k validation videos spanning 157 classes in a multi-label classification setting. Following the same experimental setups as~\cite{fan2020pyslowfast}, we fine-tune the models pretrained on Kinetics-400. For $N=5$, we train the models for 24 epochs using a step-wise learning rate schedule. The initial learning rate is set as 0.1875 per node (mini-batch size is 16 per node) and is reduced by a factor of 10 at 16 and 20 epochs. We use linear warm-up for the first 2 epochs and weight decay of $10^{-4}$. We use temporal max-pool prediction scores for inference~\cite{feichtenhofer2019slowfast}.

\subsection{Something-Something-V1}
Something-Something-V1~\cite{goyal2017something} includes about 86K training and 11.5k validation videos covering 174 classes. Following~\cite{fan2020pyslowfast}, we fine-tune the models pretrained on Kinetics-400. For $N=5$, we train the models for 11 epochs using a step-wise learning rate schedule. The initial learning rate is 0.03 per node (mini-batch size is 16 per node) and is reduced by a factor of 10 at 7 and 9 epochs. Weight decay is set as $10^{-6}$.

\subsection{AVA}
We use the same experimental setup as SlowFast~\cite{feichtenhofer2019slowfast} for action detection on AVA. Specifically, the detector is similar to Faster R-CNN~\cite{ren2015faster} with modifications adapted to the SlowFast backbone.  Region-of-interest (RoI) features are extracted at the last feature map of res$_5$ by extending a 2D proposal at a frame along the time axis into a 3D RoI, followed by application of frame-wise RoIAlign~\cite{he2017mask}, temporal global average pooling and spatial max pooling.
Following AIA~\cite{tang2020asynchronous}, we use a two-layer MLP to process the RoI features, followed by a per-class, sigmoid classifier optimized with focal loss~\cite{lin2017focal} for multi-label prediction. Note that we do not use optical flow, audio or objection detection predictions that are used in prior work~\cite{Sun2018ACRN,xiao2020avslowfast,tang2020asynchronous}.

For the training on AVA, we fine-tune the models pretrained on Kinetics. We use both ground-truth boxes and pre-computed human proposals with scores at least 0.9 for training. The human proposals are provided by~\cite{fan2020pyslowfast}. For $N=9$, we train the model for 18 epochs using the step-wise learning rate schedule. The initial learning rate is 0.0018 per node (mini-batch size is 8 per node) and is reduced by a factor of 10 at 11 and 15 epochs. We use weight decay of $10^{-7}$. At test time we generate detection results on human proposals with scores $\geq 0.8$.